\pdfoutput=1
\documentclass{article}
\usepackage[utf8]{inputenc}
\usepackage[authoryear]{natbib}
\usepackage{url}
\usepackage{graphics,graphicx,color}
\usepackage{todonotes}
\usepackage{geometry}
\usepackage{pdflscape}
\usepackage[margin=3cm]{caption}
\usepackage{lingmacros}

\title{Pilot study for the COST Action \emph{Reassembling the Republic of Letters}: language-driven network analysis of letters from the Hartlib's Papers}

\author{Barbara McGillivray, Federico Sangati}
\date{4 April 2016}

\begin{document}

\maketitle

\section{Context and Goals}

The applications of Social Network Analysis \citep{scott_2013} to literary and historical texts have attracted a growing interest in the scholarly community as powerful tools to investigate social structures. At the same time, the increased access to large amounts of digitized historical texts and the availability of corpus tools and computational methods for analysing those data in automatic ways offer new answers to humanistic research questions. Over the past decades, an increasing number of academic projects have focused on the role played by corpora in historical investigations, and several studies have shown that historical corpora contribute effectively to the progress of historical research (cf. e.g. \citealt{knooihuizena-dediub_2012, piotrowski_2012}). 

The present report summarizes an exploratory study which we carried out in the context of the COST Action  IS1310 \textit{Reassembling the Republic of Letters, 1500--1800}, and which is relevant to the activities of Working Group 3 ``Texts and Topics'' and Working Group 2 ``People and Networks''. In this study we investigated the use of Natural Language Processing (NLP) and Network Text Analysis~\citep{popping1997network, diesner2004using} on a small sample of seventeenth-century letters selected from \emph{Hartlib Papers}, whose records are in one of the catalogues of Early Modern Letters Online (EMLO),\footnote{\url{http://emlo.bodleian.ox.ac.uk/blog/?catalogue=samuel-hartlib}}
and whose online edition is available on the website of the Humanities Research Institute at the University of Sheffield.\footnote{ \url{http://www.hrionline.ac.uk/hartlib/}}

We will outline the NLP pipeline used to automatically process the texts into a network representation following the approach by~\citet{NLE:9479659, vandeCamp:2011}, in order to identify the texts' ``narrative centrality'', i.e. the most central entities in the texts, and the relations between them. 




Network Text Analysis is typically applied to a large quantity of text, hence our goal is not to provide a complete analysis of the letters under investigation. We will instead aim to make an initial assessment of the validity of this approach, to suggest how it can scale up to a much larger set of letters, and to define which infrastructure would be needed to extend this process to a potentially multilingual  historical corpus of epistolary texts.

\section{Preprocessing Steps}

In our study we have worked on the following 13 texts,\footnote{These are all letters, apart from number 7, which is a summary text written by Samuel Hartlib.} selected from the 
archive of the Hartlib Papers:\footnote{The letters are available on the website  \url{http://www.hrionline.ac.uk/hartlib/browse}. We refer to the letters with the last name of the sender, followed by the last name of the addressee, and the date of attribution of the text.} 
\begin{enumerate}
\item Dury -- Hartlib (1628?)
\item Dury -- Hartlib (1632)
\item Dury -- Hartlib (1661)
\item Dury -- Roe (1637)
\item Dury -- St Amand (1637)
\item Dury -- Waller (1646)
\item Hartlib (1631--1633)
\item Hartlib -- Davenant (1640)
\item Hartlib -- Dury (1630)
\item Hartlib -- Pell (1657)
\item Hartlib -- Robartes (1640)
\item Hartlib -- Worthington (1659)
\item Hartlib -- Worthington (1660).
\end{enumerate}

We have chosen these texts because they span over the chronological range of the Hartlib Papers, and they cover a relatively wide range of addressees. Moreover, these texts are written in English\footnote{This is with the exception of Dury-Hartlib (1628), which contains German text in its final part. We have excluded this part from the manual syntactic analysis described in section \ref{sec:building_networks}.}; this language has the largest number of resources and NLP tools, even considering historical varieties of modern languages, and therefore provided the best conditions for the linguistic processing. 

In the rest of this section we describe the \emph{letters' acquisition} procedure,  the general \emph{NLP pre-processing steps}, and the \emph{NLP tools} we have adopted to prepare the text for the \emph{Network Text Analysis} described in section \ref{sec:building_networks}.

\subsection{Letters' Acquisition}

Although all letters were digitized and transcribed, we had to apply some manual polishing to the text (e.g. removing transcription notes, formatting tags, etc.). This procedure would not be trivial to do automatically, because the text formatting is not consistent across all sources. Moreover the letters had to be imported manually one by one from the website. A much simpler alternative procedure, which would be paramount for a larger study, is to obtain access to the raw textual data of the letters as stored in the database.

\subsection{Pre-processing Steps}

We have applied the following five textual pre-processing steps to each of the acquired letters:

\paragraph{1) Sentence splitting} The typical contextual unit of reference in NLP analysis is a sentence, and therefore sentence boundaries need to be detected. This is a rather simple procedure whereby language-specific rules are normally used to decide in which cases certain punctuation marks (e.g., full stops, question marks, etc.) identify the end of the sentence.\footnote{The most common exception in which a full stop is not a sentence boundary is when it is used for abbreviations, e.g., Mr., Mrs., etc.}

\paragraph{2) Tokenization} The basic units of reference for automatic textual analyses are word tokens. These are identified by language-specific rules and separated by adjacent elements such as punctuation marks or other word tokens in agglutinative languages (e.g., the German compound \textit{Computerlinguistik}  `computational linguistics' can be tokenized as two tokens, \textit{Computer} and \textit{linguistik}).

\paragraph{3) Part of Speech (PoS) Tagging} Each token is analyzed and assigned with a specific category depending on its syntactic role (e.g., verb, noun, adjective, adverb, etc.)

\paragraph{4) Lemmatization} In order to reduce data sparsity, automatic textual analysis often resorts in lemmatizing the text, i.e., turning each inflected or variant word-token form into its basic form (e.g., eating $\to$ eat). 

\paragraph{5) Dependency Parsing} The final step is to derive the full syntactic structure of each sentence, e.g., in terms of its subjects, predicates, and objects. This is important in order to identify the argument structure of a sentence, which can be used to derive the actions, actors, and patients (\emph{who does what to whom}) in the sentence. This is preliminary to a full semantic analysis, which falls outside the scope of this study.

\subsection{NLP Tools}

We have adopted and compared two different NLP processing tools to analyze the letters. 

\begin{figure}
\centering
\includegraphics[width=0.9\linewidth, trim = 15mm 65mm 15mm 65mm, clip]{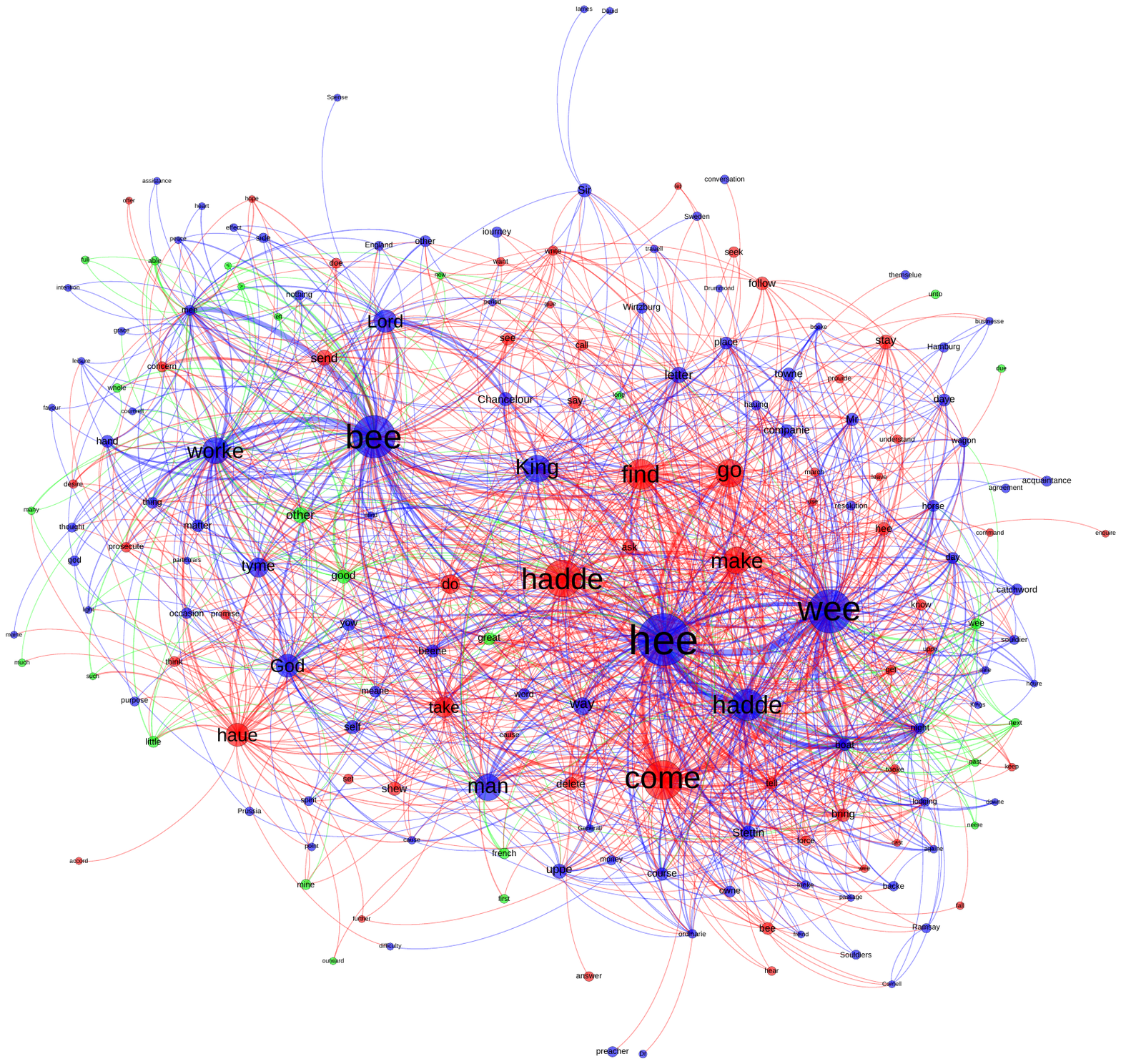}
\caption{Network of the 1628 letter from Dury to Hartlib from Stanford preprocessing with co-occurences analysis and mean-based for nodes and arcs.}
\label{stanford}
\end{figure}

\paragraph{1) Stanford Core NLP Tools} This is one of the most complete state-of-the art NLP libraries \citep{manning-EtAl:2014:P14-5}, which  implements all five pre-processing steps for several modern languages. Since it has no ready model for Early Modern English, we used the model for modern English.\footnote{This has caused several PoS tagging errors, such as `bee' in figure~\ref{stanford} classified as a noun instead of a form of the verb \textit{be}.} It is however in principle possible to train new language models (and therefore a model for Early Modern English), provided enough annotated materials for such languages.

\paragraph{2) MorphAdorner} This tool \citep{morphadorner} is one of the most commonly used tools for NLP processing of historical English. 
It requires the text to be in TEI (Text Encoding Initiative) format; therefore we manually added a TEI header for epistolary texts to each text. This step can be easily automated. MorphAdorner performs all the pre-processing steps described above except for the first (sentence splitting) and the last one (dependency parsing). The sentence splitting can be automated. 

In the next section we describe an alternative method to identify the basic argument structure of a sentence without dependency parsing; this method is based on subject-verb-object triplets obtained from their relative positions in the sentence. We will also describe how we constructed the networks.

\section{Building the networks}\label{sec:building_networks}

In this section we describe the procedure we followed to build a network (or graph) from a preprocessed text. A selected set of networks from the texts under investigation is reported in the Appendix A. These were built with automatic codes that use the Gephi library \citep{ICWSM09154} for rendering the networks in a graphical mode.

The basic elements in our networks are lemmatized word tokens represented as circles (nodes), with specific colors depending on their PoS categories: red for verbs, blue for nouns and green for adjectives. The lines (arcs) connecting two nodes represent a specific relation between them.

All code is open source and available at \url{https://github.com/kercos/DH_Code}.

\subsection{Word relations}
We illustrate two basic methodologies for building the networks: one based on word \emph{co-occurrences} and the other based on their \emph{syntactic relations}.

\paragraph{1) Co-occurrences}
The simplest way to build a network from a text is to rely on co-occurrence information, that is two word tokens (e.g. a noun and a verb) are connected if they co-occur in the same textual context.  Additionally, we want to keep track of the frequency of these connections to distinguish word-token pairs which co-occur more or less often. 
We have performed an automatic co-occurrence extraction using our own code, starting both from the Stanford and the MorphAdorner preprocessed texts. In the current analysis we consider the sentence as the contextual unit to extract co-occurrences. A common alternative is to restrict the contextual region to a window of a specific number of words (typically 4 or 5). 

\paragraph{2) Syntactic relations}
A more refined way to represent connections between entities in a given text is to visualize their syntactic relations \citep{tanev-magnini_2008,McGillivray:2008:SSC:1627328.1627335}. In the current study we focused on a subject-verb-object triplet representation and extracted such triplets by hand for one letter (see section \ref{sec:analysis_dury-harlib-1628}). In order to automate this step we would need a dependency parsing processing. The Stanford NLP Tools provide a dependency parser for modern English. For what concerns historical English, it is possible to develop a parser based on manually annotated texts, and some research has already been done in this direction, as summarized in \citet {piotrowski_2012}.

Since the texts pre-processed with MorphAdorner lacked the syntactic information required to extract the subject-verb-object triplets, we devised a workaround to obtain a similar representation based on the typical word-order of English: for every verb in the sentence we identified the closest noun on its left as the candidate subject (with a maximum distance of four words), and the closest noun on its right as the candidate object (with a maximum distance of four words).\footnote{As we explain in section \ref{sec:final_remarks}, this workaround is unlikely to work well for languages with a freer word order, such as Latin.} 

For example, let us consider the following sentence, from the letter from John Dury to Samuel Hartlib (1628):

\enumsentence{I begin to \textbf{shew} what \textbf{prudency} \& care a \textbf{Tutour} must \textbf{vse} to \textbf{move} little \textbf{Children} [\ldots]}\label{ex_trip}

From \ref{ex_trip}, we extracted the following pairs (the verb and noun lemmas are listed):

\begin{itemize}
    \item \emph{show} - \emph{prudency}
    \item \emph{Tutour} - \emph{use}
    \item \emph{move} - \emph{Children}
\end{itemize} 

In the list above, we note that we did not extract pronouns, but we will consider pronouns in the manual analysis reported on in section \ref{sec:analysis_dury-harlib-1628}; moreover, instead of triplets, we were only able to extract pairs of candidate subjects and verbs or verbs and candidate objects. Finally, note that \emph{prudency} is not a direct object of \emph{show} because of the indirect clause following this verb, showing that the context-based triplets do not perfectly reflect the syntactic relationships between items. In section \ref{sec:analysis_dury-harlib-1628}  we will suggest how this can be improved thanks to a manual syntactic analysis, which can be automated.

\subsection{Pruning the networks}

The number of nodes and connections tends to grow extremely large with the size of the text. It is therefore necessary to show only the most representative ones, i.e., those occurring more frequently. This is accomplished by removing (pruning) less frequent nodes and connections, which tend to also be the less reliable ones. As a matter of fact, although the methodology is prone to detect a number of erroneous connections, in a very large text these errors will tend to have a low frequency.

We have adopted two basic pruning strategies:
\begin{description}
\item[Number-based:] we select only nodes and arcs whose frequency is above a predefined threshold (e.g., FREQ $>$ 1, selects only elements with frequency greater than 1).
\item[Mean-based:] we select only nodes and arcs whose frequency is above the MEAN of the respective frequency distribution plus a certain number of standard deviations (e.g., MEAN + 2SD, selects only elements with frequency greater than the mean plus two standard deviations).

\end{description}

\section{Analysis}\label{sec_analysis}

\begin{figure}
\centering
\includegraphics[width=0.9\linewidth, trim = 15mm 65mm 15mm 65mm, clip]{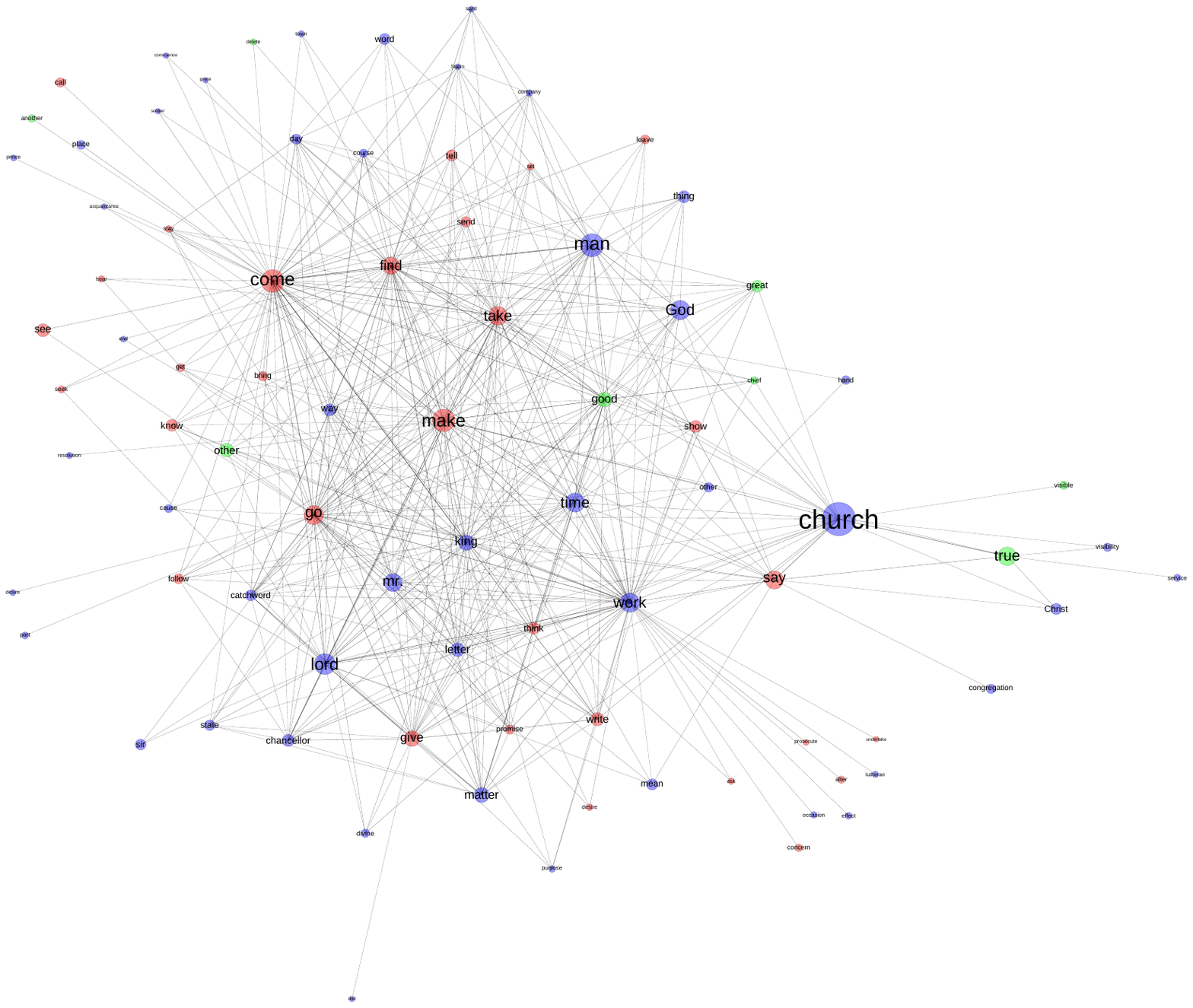}
\caption{Network of the 1628 letter from Dury to Hartlib, from MorphAdorner preprocessing with co-occurences analysis and mean-based pruning (MEAN+1SD for nodes and MEAN+2SD  for arcs).}\label{mode_1_2cooccorrenze}
\end{figure}

\begin{figure}
\centering
\includegraphics[width=0.9\linewidth, trim = 15mm 55mm 15mm 55mm, clip]{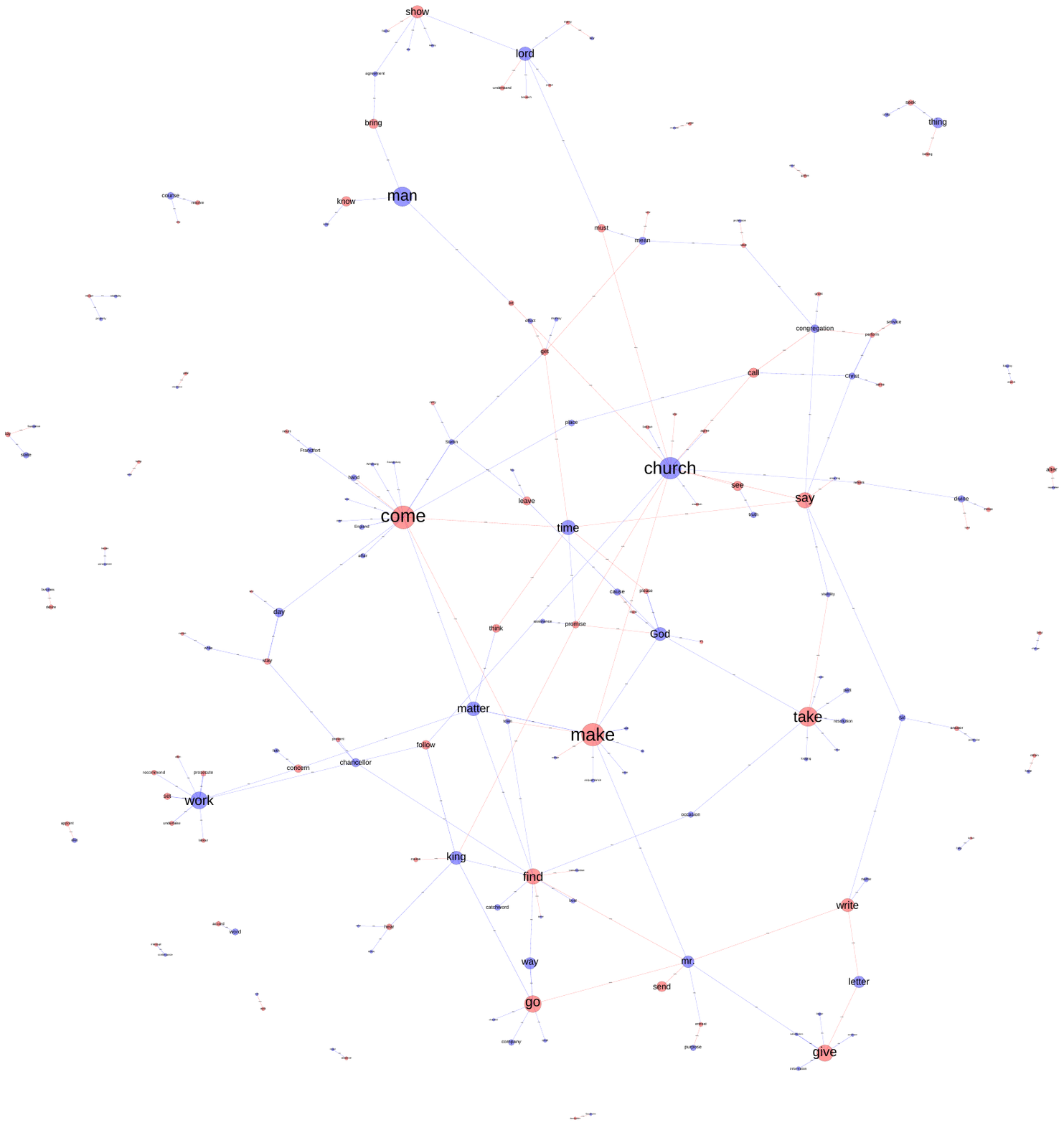}
\caption{Network of all 13 letters from MorphAdorner preprocessing with triplet analysis and number-based pruning (FREQ$>$1 for nodes and FREQ$>$2 for arcs).}
\label{mode_1_1_DEPtriple}
\end{figure}

\begin{figure}
\centering
\includegraphics[width=0.9\linewidth, trim = 15mm 65mm 15mm 65mm, clip]{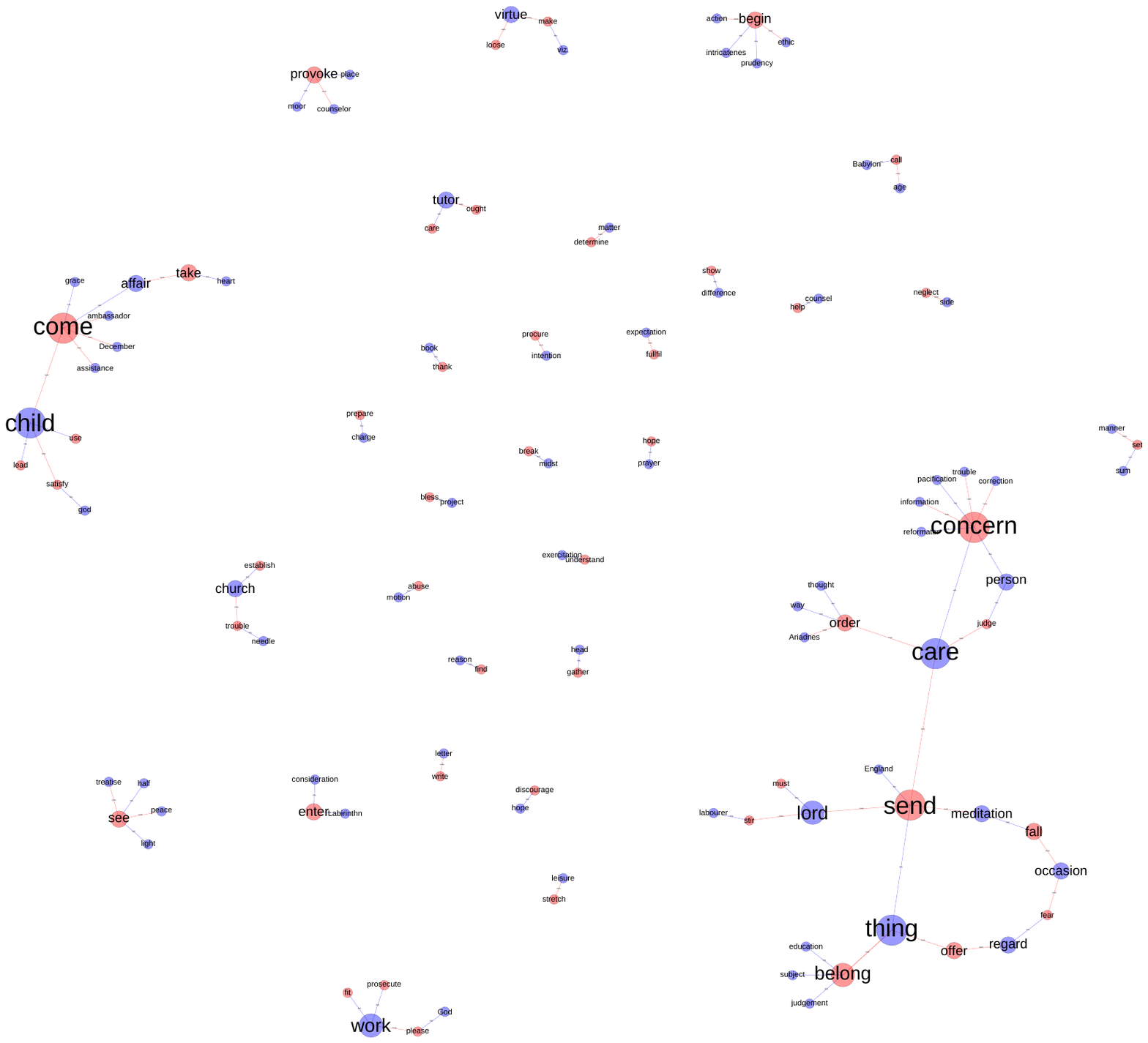}
\caption{Unpruned network of the 1628 letter from Dury to Hartlib; the preprocessing was performed with MorphAdorner and was followed by an automatic extraction of triplets.}
\label{1_manuale}
\end{figure}

\begin{figure}
\centering
\includegraphics[width=0.9\linewidth, trim = 15mm 65mm 15mm 65mm, clip]{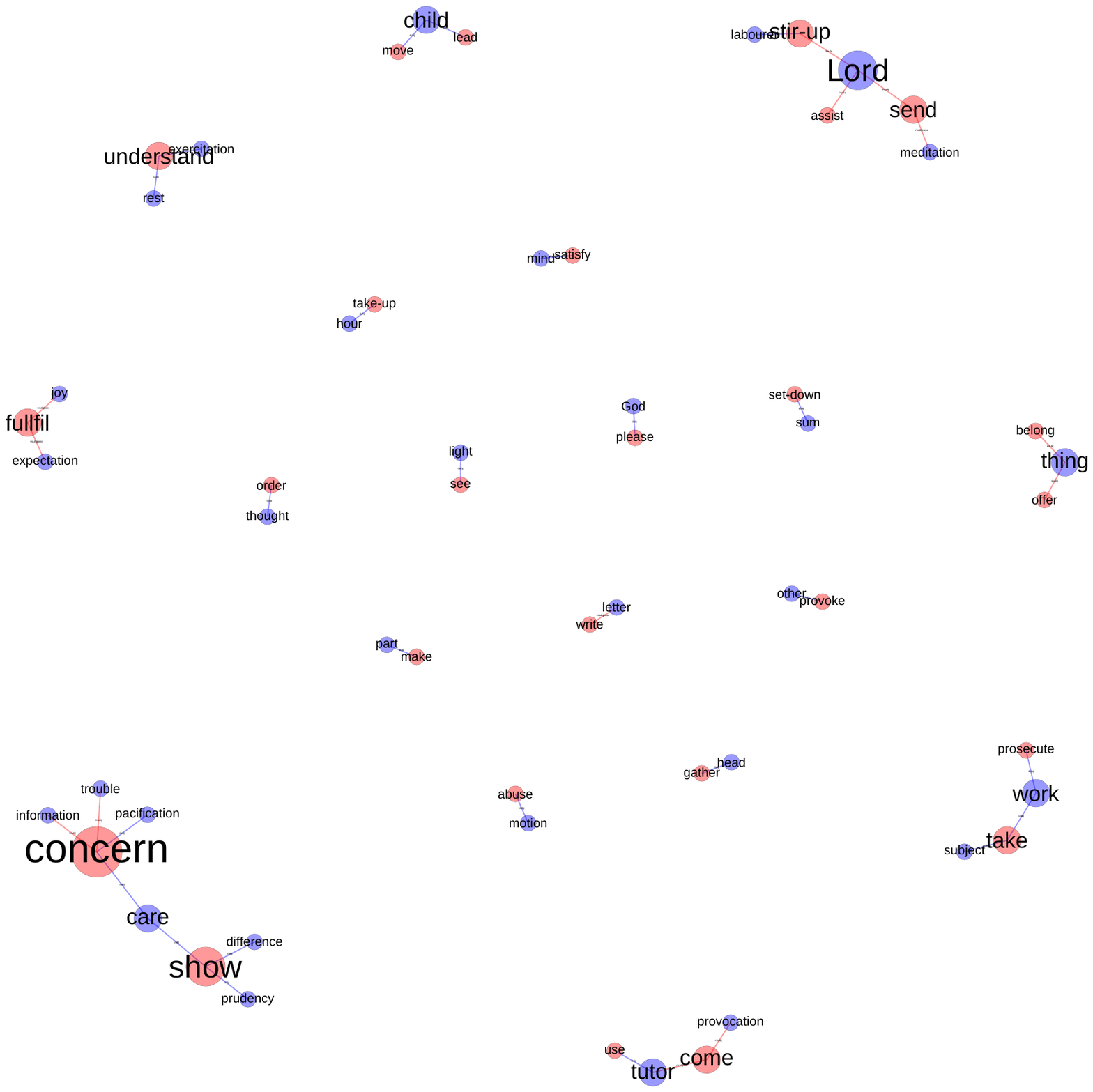}
\caption{Unpruned network of the 1628 letter from Dury to Hartlib; the preprocessing was performed with MorphAdorner and was followed by a manual extraction of triplets.}
\label{2_estrazione manuale}
\end{figure}

\begin{figure}
\centering
\includegraphics[width=0.9\linewidth, trim = 15mm 50mm 15mm 50mm, clip]{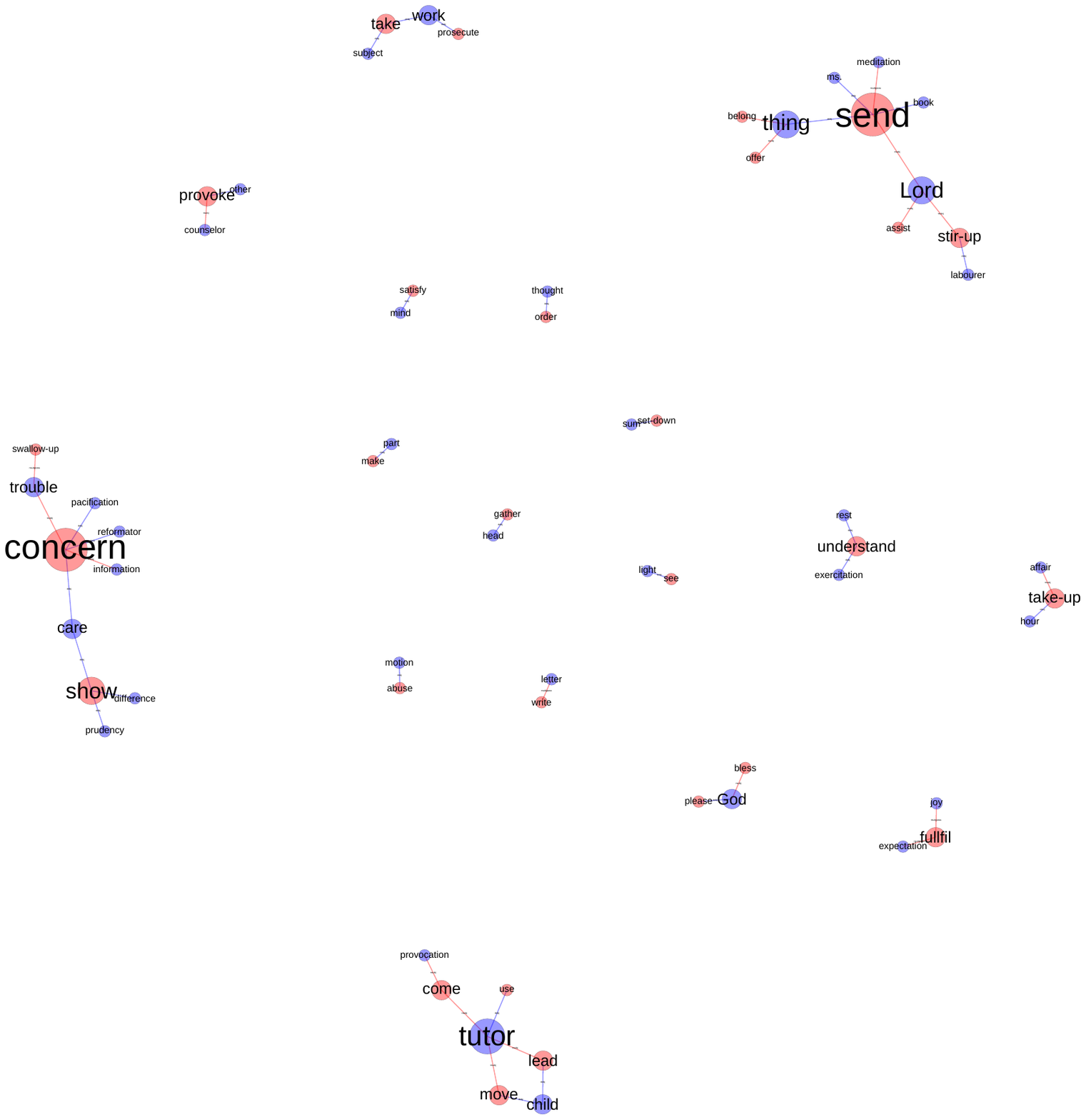}
\caption{Unpruned network of the 1628 letter from Dury to Hartlib; the preprocessing was performed with MorphAdorner and was followed by a manual extraction of triplets and anaphora resolution.}
\label{3_estrazionemanuale_anafora}
\end{figure}

As detailed in section \ref{sec:building_networks}, we created a number of different networks, showing the various steps of our approach. In this section we will focus on two groups: 

\begin{itemize}
\item networks relative to the collection of all 13 letters (Figures \ref{mode_2_2cooccorrenze} and \ref{mode_1_1_DEPtriple}) 
\item networks relative to the letter sent by John Dury to Samuel Hartlib around 1628, and available at \url{http://www.hrionline.ac.uk/hartlib/browse.jsp?id=1%2F12%2F1a-b} (Figures \ref{stanford}, \ref{mode_1_2cooccorrenze}, \ref{1_manuale}, \ref{2_estrazione manuale}, and \ref{3_estrazionemanuale_anafora})
\end{itemize} 

Given the small size of the corpus considered, we do not provide a quantitative analysis of the data. Therefore, we will limit ourselves to general observations and focus on the methodological implications of our approach and its potential for broader applications.  

\subsection{Networks from the collection of letters}

We derived the first group of networks in a fully automatic way by first pre-processing the letters (tokenization, sentence segmentation, and lemmatization), and then by automatically extracting co-occurrence patterns. 

\begin{figure}
\centering
\includegraphics[width=0.9\linewidth, trim = 10mm 30mm 10mm 30mm, clip]{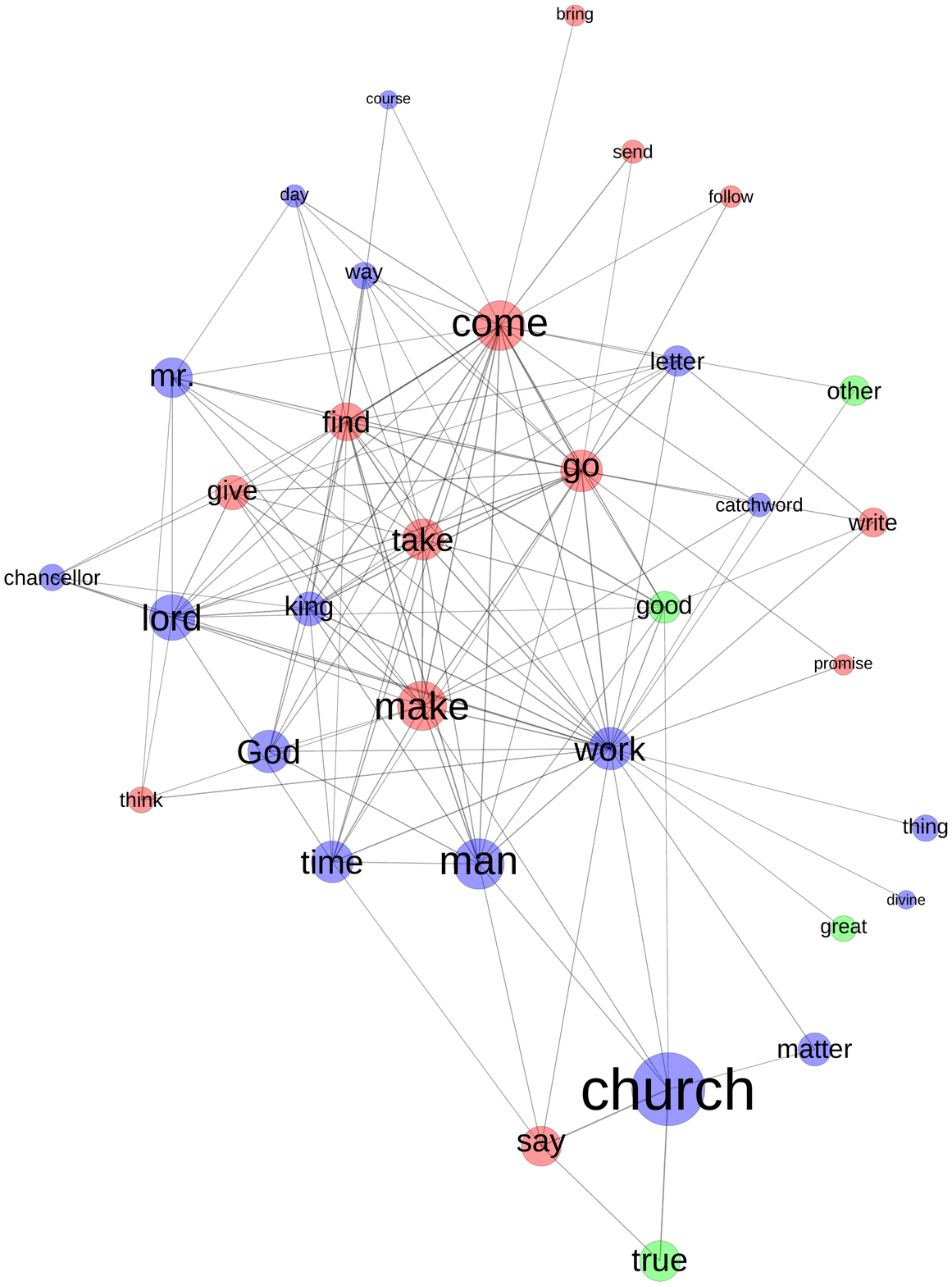}
\caption{Network of all 13 letters from MorphAdorner preprocessing with co-occurences analysis and mean-based pruning (MEAN+2SD for nodes and MEAN+2SD for arcs).}\label{mode_2_2cooccorrenze}
\end{figure}

The nodes in the network in Figure  \ref{mode_2_2cooccorrenze} correspond to the lemmas of nouns (blue nodes), verbs (red nodes), and adjectives (green nodes) occurring in the letters, and their size is proportional to the frequency of the lemmas in the corpus; if two nodes are connected, it means that they occur in the same sentence. As the sentences in the letters are often long, these networks display the most frequent entities  (nouns) and actions (verbs) mentioned in the letters. 

Figure \ref{mode_2_2cooccorrenze} summarizes the main topics that the letters are concerned with: \textit{church}, \textit{man}, \textit{God}, \textit{Lord}, \textit{time}, and \textit{work} regarding nouns, and \textit{come}, \textit{make}, \textit{take}, and \textit{find} regarding verbs.
By contrast, Figure \ref{mode_1_1_DEPtriple} was obtained by considering a narrower context of co-occurrences for verbs and nouns, which led to results that are closer to an actor-action model.
The red edges link verbs to the nouns occurring before them in a window of four words (candidate subjects), and the blue edges link verbs to the nouns occurring after them in the same four-word window (candidate objects). This approach to detecting candidate subjects and objects is not always accurate, as we explain below. 

Let us consider the noun \textit{truth}, connected to the verb \textit{see} by a blue edge, indicating a candidate object role. In fact, \textit{truth} follows a form of \textit{see} twice in the corpus, in both cases in the letter from John Dury to Joseph St Amand (1637), available at \url{http://www.hrionline.ac.uk/hartlib/browse.jsp?id=26%2F19%2F6a-7b}:

\enumsentence{By this then wee \textbf{see} what \textbf{trueth} is [\ldots].}\label{ex1}

\enumsentence{First the Congregation it selfe is to bee \textbf{seene}: And secondly the \textbf{trueth} or the falshood of the service perfourmed to Christ in the congregation.}\label{ex2}

In \ref{ex1} the verb governs a clause introduced by \textit{what} whose subject is \textit{truth}, so \textit{trueth} is not strictly speaking the direct object of \textit{see}, even though from a semantic point of view this is not completely inaccurate. In \ref{ex2}, however, the algorithm ignores sentence boundaries marked by colons, as we only considered full stops as sentence delimiters. One simple way to avoid these kinds of errors would be to use colons to identify the clause boundaries and impose this as a constraint for the algorithm. 
In other cases the errors concern other syntactic phenomena, which are more difficult to address in absence of a full syntactic parsing. For example, in the same letter we find:

\enumsentence{Because before Luthers time the \textbf{Church} which is now \textbf{called} the Protestant Church had no being nor visibilitie [\ldots].}\label{dury-stamand-ex1}

In this case \textit{Church} is the subject of a passive form of \textit{call}, and therefore, even though it occurs before the verb, it is not a subject.
In order to partially remedy these problems, we have included syntactic information for one of the letters, as we show in the next section.

\subsection{Networks from the 1628 letter from Dury to Hartlib}\label{sec:analysis_dury-harlib-1628}

Figure \ref{stanford} shows the network for the 1628 letter from Dury to Hartlib, obtained from data processed with the Stanford parser. We can notice that the pronouns \emph{wee} `we' and \emph{hee} `he' are incorrectly lemmatized and tagged as nouns, and that the verb \emph{bee} `be' is incorrectly lemmatized and tagged as a noun.

Figure \ref{mode_1_2cooccorrenze} shows the network derived from the letter preprocessed with MorphAdorner; the edges correspond to the co-occurences analysis and the network was pruned based onmean-based pruning (MEAN+ the 1SD for nodes and MEAN+2SD  for arcs)

Figure \ref{1_manuale} contains the network with the context-based definitions of candidate subjects and objects of the verbs occurring in the letter from John Dury to Samuel Hartlib (1628), while Figure \ref{2_estrazione manuale} displays manually annotated subjects and objects and their verbs. As we can see from the comparison of the two figures, the latter is definitely a more accurate representation of the entities and actions mentioned in the letter. While keeping in mind that this is the analysis of a single letter and that we need to be cautious in any generalization, we will make some general remarks that support the validity and potential of this approach. 

In addition to some known collocations\footnote{A collocation is a sequence of two or more words that tend to occur often together.} (e.g. \textit{see light}, \textit{please God}), we can identify active and passive entities from the point of view of their syntactic role in the sentences. For example, we observe that the noun \textit{child} is object of the verbs \textit{move} and \textit{lead}, suggesting a patient role. This may be opposed to the active role of \textit{tutor} (subject of \textit{come}).
The \textit{Lord} predominantly appears as an actor (subject of \textit{assist}, \textit{stir up}, and \textit{send}), possibly suggesting the idea of an interventionist God.
Coming to inanimate entities, thoughts appear in need to be ordered (\textit{thought} is the object of \textit{order}). Moreover, topics of concern seem to be the prevention of negative outcomes, as suggested by the nouns associated with the verb \textit{concern}, like \textit{pacification}, \textit{(pastoral) care}, and \textit{trouble}.

Figure \ref{3_estrazionemanuale_anafora} is derived from an additional anaphora resolution step,\footnote{In a linguistic context, \emph{anaphora resolution} refers to the resolution of an expression based on another expression occurring before or after it (its antecedent or postcedent, respectively).} which contributes to making the analysis richer. For example, we notice that now the nodes \textit{child} and \textit{tutor} are connected because \textit{tutor} is the subject of \textit{move} and \textit{lead}, which have \textit{child} as their object. 
Let us look at the relevant passages:

\enumsentence{I begin to shew what prudency \& care a \textbf{Tutour} must vse to \textbf{move} little \textbf{Children} that are vncapable of the Precepts of Christianity to a Custome of naturall vertues [\ldots]}\label{ex_dury-hartlib_1}

\enumsentence{[\ldots] seeking to enter into a particular consideracion of the whole duty of a \textbf{Tutour} how hee ought to bee fitted \& prepared for the Charge \& what hee ought to doe to \textbf{leade} a \textbf{Child} from his infancy as it were by the hand through an insensible Custome of well doeing vnto a perfect degree of all vertues}\label{ex_dury-hartlib_2}

As the excerpts above attest, the tutor is the entity performing the action of moving and leading, respectively in \ref{ex_dury-hartlib_1} and in \ref{ex_dury-hartlib_2}. In \ref{ex_dury-hartlib_1} specifically, after the anaphora resolution step, the pronoun \emph{hee} `he' is resolved to refer to \emph{Tutour} `tutor'.

Of course, only a systematic quantitative analysis on a larger scale would be able to confirm the preliminary observations done here. However, we have shown that the networks are able to provide some insights into the content of the letters, as we summarize in the next section.

\section{Final Remarks}\label{sec:final_remarks}

The results we have achieved show that the approach is promising and, if extended in its scope, can lead to positive results for historical research on correspondence texts. We have also shown that methodologies developed to analyse contemporary texts have the potential to be successfully applied in a historical context, with specific adjustments.

Since the nature of this pilot study was methodological and exploratory, we have focused the analysis on a limited set of letters. However, the automatic procedures we have followed can be applied to a significantly higher number of letters. In fact, it is on large datasets that certain patterns can be detected and analyzed statistically, which is one of the strengths of computational approaches such as the present one.

\subsection{Possible extensions}

In addition to applying the processing and analysis to a larger set of letters, this study can be extended in a number of directions, as we outline below.

\begin{description}
\item[Languages] This pilot study focused on English. However, thanks to the automatic procedures followed, it can in principle be applied to data in other languages as well. This would suit the high degree of multilinguality in the Hartlib's Papers well, and capture possible interesting associations between the semantic content and the languages used. Nevertheless, the specific features of the languages might require adjustments in the syntactic processing and extraction of triplets.\footnote{For example, as the word order in Latin is freer than in English and Latin morphology is richer, the extraction of triplets based on the nouns occurring before/after verbs as candidate subjects/objects is unlikely to lead to good results. By contrast, constraints on the morphological case of the nouns (e.g. nominative for subjects and accusative for objects) and morphological agreement of the verb with the candidate subject noun are more promising, lacking a full syntactic processing. Similar arguments hold for morphologically richer languages like German or Italian}.
\item[Preprocessing]A number of steps could increase the accuracy of the preprocessing steps. 
\begin{itemize}
    \item Relying on syntactically parsed texts would make the subject-verb-object triples more accurate.
    \item Labelling nouns and verbs according to their semantic classes (such as \emph{persons} and \emph{vehicles} for nouns and \emph{communication} and \emph{motion} for verbs, just to give a few examples) would allow us to group the triplets in larger categories and detect possible patterns in larger networks.
    \item Performing the anaphora resolution automatically would enrich the analysis, as shown in section \ref{sec_analysis}.
    \item Following the approach presented in \citet{Trampus11}, rather than focusing on subject-verb-object triplets, we could extract full event patterns from the letters and therefore derive semantic graphs, where nodes represent actors and edges represent actions.
\end{itemize}
\item[Evaluation]A systematic evaluation of the annotation of the texts would be necessary to assess the quality of the data from which the networks were built. This can be done by comparing the automatically extracted triples with a manually created gold standard.
\item[Network analysis]A systematic and quantitative analysis of the networks based on centrality measures (such as in-degree and out-degree measures) would highlight particularly active actors and actions, as well as connections between them.
Further, replacing static networks with dynamic networks extracted from the full epistolary corpus would help to identify the change in importance of actors over time, as outlined in \citet{agarwal-EtAl:2012}.
\item[Further analysis] It is possible to combine the linguistic features explored in this study with the metadata of the letters (which capture historically relevant information, such as date, location, sender, and addressee), as well as other text metadata (e.g. length, structure, complexity of letter). This has the potential to offer new insights into the context, content, and structure of the letters, and support further research on this material.
\end{description}

\section{Networks}

\newpage

\newgeometry{margin=0.1cm}
\thispagestyle{empty} 

\newpage

\newgeometry{margin=0.1cm}
\thispagestyle{empty} 

\newpage

\newgeometry{margin=0.1cm}
\thispagestyle{empty} 

\newpage

\newgeometry{margin=0.1cm}
\thispagestyle{empty} 



\newpage

\newgeometry{margin=0.1cm}
\thispagestyle{empty} 

\newpage

\newgeometry{margin=0.1cm}
\thispagestyle{empty} 

\newpage

\newgeometry{margin=0.1cm}
\thispagestyle{empty} 
\restoregeometry

\newpage
\appendix 


\bibliographystyle{apa}
\bibliography{main.bib}

\end{document}